\documentclass{article}
\usepackage{ijcai26}

\usepackage{times}
\usepackage{soul}
\usepackage{url}
\usepackage[hidelinks]{hyperref}
\usepackage[utf8]{inputenc}
\usepackage[small]{caption}
\usepackage{graphicx}
\usepackage{amsmath}
\usepackage{amsthm}
\usepackage{booktabs}
\usepackage{algorithm}
\usepackage{algorithmic}
\usepackage[switch]{lineno}
\usepackage{times}
\usepackage{soul}
\usepackage{url}
\usepackage[hidelinks]{hyperref}
\usepackage[utf8]{inputenc}
\usepackage{multirow}
\usepackage{color} 
\usepackage{amssymb}
\usepackage{xspace}
\usepackage{enumitem}

\newcommand{\ie}{\emph{i.e.,}\xspace}

\title{FreSH: Frequency-Segmented Hierarchical Multi-Expert Framework \\for Multivariate Time Series Classification}

\author{
Pingping Liu$^{1}$,
Muyao Wang$^{1}$,
Zijian Zhang$^{1,*}$,
Tongshun Zhang$^{1}$,
Hao Miao$^{2}$,
Guorui Xie$^{3}$,
Qingliang Li$^{4}$,
Qiuzhan Zhou$^{1}$
\affiliations
$^{1}$Jilin University\\
$^{2}$Hong Kong Polytechnic University\\
$^{3}$Pengcheng Laboratory\\
$^{4}$Changchun Normal University
\emails
liupp@jlu.edu.cn,
wangmy24@mails.jlu.edu.cn,
zhangzijian@jlu.edu.cn,
tszhang23@mails.jlu.edu.cn,
hao.miao@polyu.edu.hk,
xiegrr@gmail.com,
liqingliang@ccsfu.edu.cn,
zhouqz@jlu.edu.cn
}

\begin{document}

\maketitle

\begin{abstract}
Multivariate Time Series Classification (MTSC) demands models that can effectively capture complex temporal patterns across multiple scales while remaining computationally efficient. However, existing approaches generally struggle to reconcile fine-grained representation learning, especially under class imbalance and real-world constraints. In this paper, we present FreSH, a Frequency-Segmented Hierarchical Multi-Expert Framework designed to address these challenges. FreSH introduces a new perspective for MTSC by enabling adaptive, multi-scale analysis of temporal signals, allowing different aspects of the data to be modeled in a complementary and coordinated manner. By combining localized specialization with holistic context modeling, FreSH achieves strong representational capacity without incurring excessive computational overhead. An adaptive fusion strategy further enhances flexibility, enabling the model to dynamically emphasize the most informative components of the input. In addition, we incorporate a more robust optimization objective that improves learning stability across varying sample difficulties and class distributions. Extensive evaluations on 30 UEA benchmark datasets and real-world vibration data demonstrate that FreSH consistently outperforms state-of-the-art methods in classification accuracy, while substantially reducing model size and efficiency. 
The implementation code is publicly available at \url{https://github.com/Wangmy2120/FreSH00}.

\end{abstract}

\section{Introduction}

\begin{figure}[t]
\centering
\includegraphics[width=1.0\columnwidth]{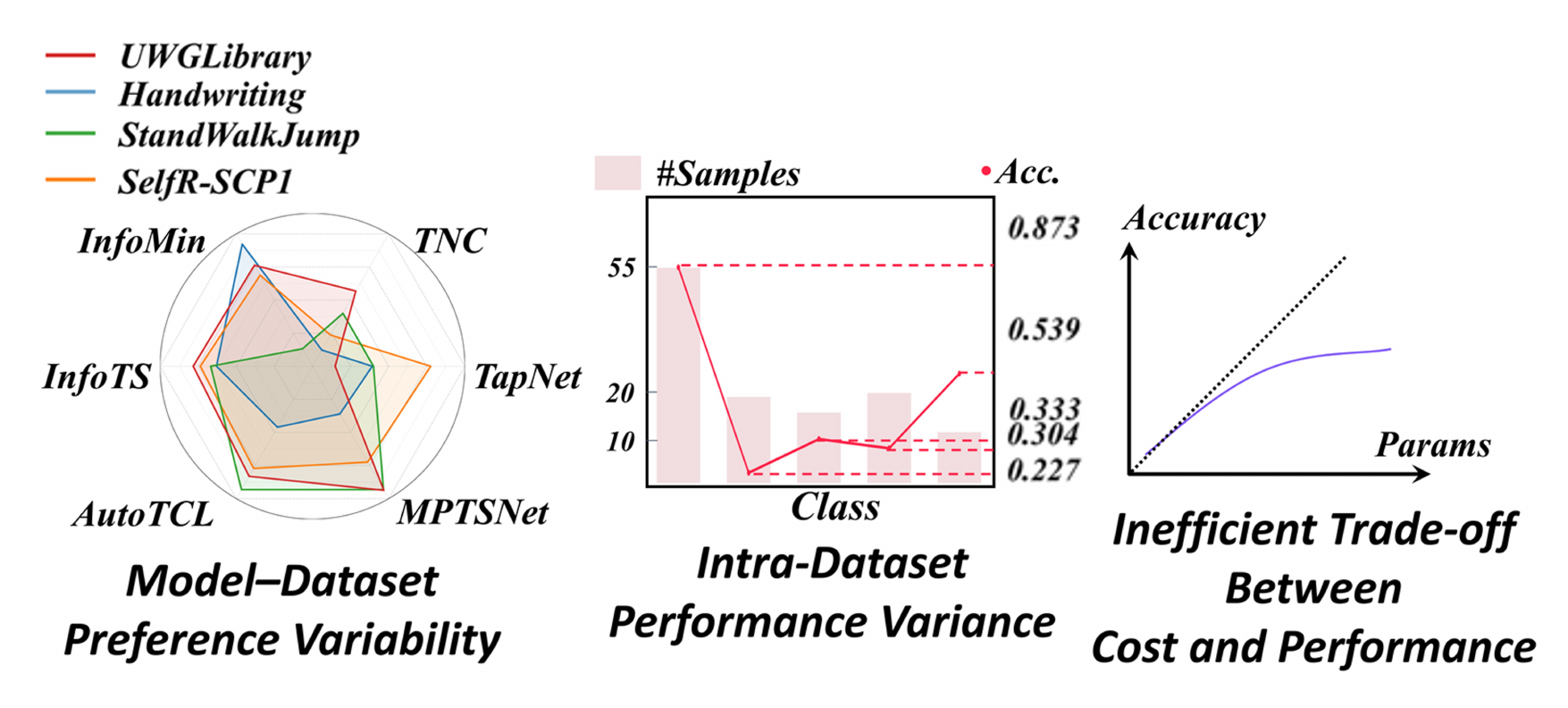} 
\caption{
Some models in MTSC cannot adapt to the diverse temporal patterns of different datasets, resulting in performance differences between datasets and across different categories. 
}
\label{fig1}
\end{figure}
Multivariate time series classification has attracted significant attention due to its broad applications in healthcare~\cite{an2023comprehensive}, industrial equipment fault diagnosis~\cite{farahani2023time}, and human action recognition~\cite{li2023human}. Accurate time series classification provides crucial support and insights for decision-makers.
However, inherent properties of time series data, such as complex dynamics, noise, and class imbalance, make MTSC a particularly challenging task~\cite{ismail2019deep}.

Traditional MTSC algorithms, such as DTW~\cite{wang2017time}, primarily rely on feature statistics or signal processing techniques. As datasets become more complex, these methods struggle to scale to modern, high-dimensional time series and fail to generalize across diverse application scenarios
~\cite{ruiz2021great}.
Recently, deep learning has emerged as the dominant paradigm for MTSC. CNN-based methods, such as OS-CNN~\cite{tang2020omni}, excel in learning spatial hierarchical features through convolutional filters but are limited in comprehensively modeling global features, often requiring additional designs to compensate for this shortcoming. RNN-based~\cite{karim2017lstm} methods face challenges in capturing long-term dependencies due to vanishing gradients. Transformer-based models ~\cite{wen2022transformers,zuo2023svp} propose to handle long-range dependencies through self-attention mechanisms but fall short in extracting local pattern features at adjacent time points.

Despite the promising progress achieved by existing methods in time series classification, several key limitations hinder their performance and practicality. 
{First}, {existing models often struggle to effectively capture the intricate, multi-scale nature of time series data.} They typically process either time-domain data directly~\cite{he2015early} or a holistic frequency-domain ~\cite{yi2023frequency}, thereby failing to distinguish and analyze the distinct information carried by different frequency bands. 
The diversity of the MTSC dataset poses challenges for existing models in balancing differences between datasets and across categories. As shown in Figure \ref{fig1}, different models exhibit significant performance variations on different types of UEA datasets, and their accuracy is markedly affected by the number of samples in different categories, with performance improvements accompanied by considerable overhead.
{Furthermore}, recent complex models, particularly those leveraging global self-attention mechanisms~\cite{zhou2021informer}, suffer from high computational costs and poor scalability, making them unsuitable for real-time applications or large-scale datasets. Simultaneously, these models lack the adaptability to specialize their processing based on the local characteristics of the data. 
{Finally}, conventional loss functions in MTSC, such as Cross-Entropy~\cite{wu2022timesnet} and Focal Loss~\cite{lin2017focal}, present a dilemma: the former is often dominated by majority classes, while the latter can over-correct for difficult samples. This makes them suboptimal for handling the joint challenges of class imbalance and varying sample difficulty, which are common in real-world time series datasets.

Motivated by these challenges, we propose a \textbf{Fre}quency-\textbf{S}egmented \textbf{H}ierarchical Multi-Expert Framework for time series classification, \ie \textbf{FreSH}. 
Our main goal is to go beyond a single-view approach by using the frequency domain and proposing a segmentation strategy that allows us to specifically analyze different frequency components. 
To fix the lack of flexibility and high costs of current models, we design an efficient and adaptive hierarchical multi-expert system. This architecture uses dedicated local experts for specific data segments while a global expert provides overall context, all within a lightweight framework that avoids complex and slow mechanisms like global attention. 
An adaptive gating mechanism fuses these outputs, dynamically weighting local frequency bands and global frequency spectrum. 
We also aim for a better optimization strategy by introducing a polynomial Loss as an alternative to standard loss functions, which we believe can create a better balance when optimizing for both easy and difficult samples, as well as majority and minority classes.
Our major contributions can be summarized as follows:

\begin{itemize}[leftmargin=*]
    \item 
    We propose a frequency-aware modeling paradigm that adaptively captures the multi-scale and multi-band characteristics of multivariate time series, offering a principled alternative to single-view time- or frequency-domain approaches for time series classification.
    \item 
    We design FreSH, a lightweight frequency-segmented hierarchical multi-expert framework that enables adaptive specialization across frequency components by integrating local and global experts, achieving effective multi-scale representation with low computational overhead.
    \item Extensive experiments on 30 UEA benchmark datasets validate the effectiveness and generalization capability of our FreSH. Our method outperforms diverse state-of-the-art baselines while achieving advanced efficiency.
\end{itemize}

\section{Methodology}

\subsection{Problem Formulation}

Let $\mathcal{X} = \{ \mathbf{X}_i\}_1^n $ represent a multivariate time series dataset, where each sample $ \mathbf{X}_i \in \mathbb{R}^{d \times l} $ represents the observations of \( d \) variables over \( l \) time steps, the goal of multivariate time series classification is to learn a classifier $f_\theta$ to accurately predict the corresponding label, \ie $  \mathbf{X}_i \in \mathbb{R}^{d \times l} \xrightarrow{f_\theta} \hat{\mathcal{Y}}_i \in \mathbb{R}^{c} $.

\subsection{Framework Overview}
\begin{figure}[!t]
\centering
\includegraphics[width=1\columnwidth]{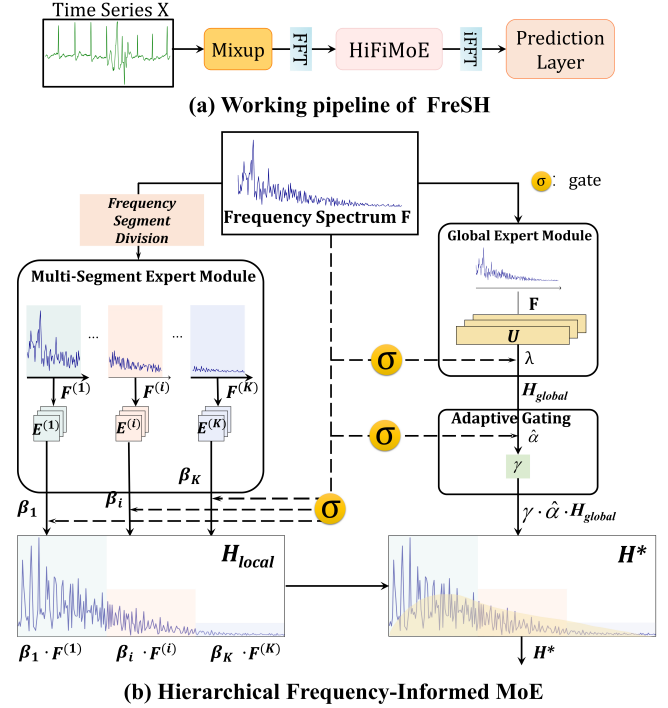} 
\caption{Framework overview of FreSH. After transforming time series into the frequency domain, the \textbf{Multi-Segment Expert Module} learns segment-wise patterns, the \textbf{Global Expert Module} captures full-spectrum dependencies, and the \textbf{Adaptive Gating Mechanism} adaptively fuses them for prediction. 
}
\label{fig2}
\end{figure}

We propose \textbf{FreSH}, a frequency-domain expert framework for multivariate time series classification, which integrates localized and global modeling within a mixture-of-experts network.
The working pipeline is shown in Figure \ref{fig2} (a). We conduct mixup for different samples to enrich the data diversity first, and then transform the data into the frequency domain and process it by a Hierarchical Frequency-Informed MoE (HiFiMoE) module, shown in Figure \ref{fig2} (b). The output is fed to the prediction layer for classification.

\subsection{Data Preprocessing}

Given an input multivariate time series sample \( \mathbf{X}_i \in \mathbb{R}^{d \times l} \), where \( d \) is the number of variables and \( l \) is the sequence length, we first conduct mixup~\cite{zhang2017mixup} to enrich data diversity.
We transform it into the frequency domain using the Fast Fourier Transform:
\begin{equation}
\mathbf{F} = \mathrm{FFT}(\mathbf{X}_i) \in \mathbb{C}^{d \times s},
\end{equation}
where $s = \left\lfloor \frac{l}{2} \right\rfloor + 1
$. This frequency-domain representation allows the model to exploit periodic patterns and frequency-specific information that are difficult to capture in the time domain. The resulting signal \( \mathbf{F} \) is then zero-padded to a fixed length \( s_{\text{padded}} \) to ensure divisibility and consistency across samples.

\subsection{Frequency Segment Division}

To capture the unevenly distributed information in the frequency domain, we divide the spectrum into multiple segments. This segmentation allows specialized experts to focus on specific bands, enabling more targeted feature learning and preparing for adaptive fusion later.
Specifically, we divide the entire frequency spectrum into \( K \) equal-length segments:
\begin{equation}
\mathbf{F} = [\mathbf{F}^{(1)}, \mathbf{F}^{(2)}, \dots, \mathbf{F}^{(K)}],
\end{equation}
where each segment \( \mathbf{F}^{(k)} \in \mathbb{C}^{d \times l_k} \) corresponds to a specific frequency band, and $l_k \times K = s_{padded}$. 
This segment division operation offers the opportunity for the specific utilization of individual frequency components that may carry unique patterns relevant for time series classification.

\subsection{Hierarchical Frequency-Informed MoE}
\subsubsection{Multi-Segment Expert Module}

For each frequency segment \( \mathbf{F}^{(k)} \), we design a dedicated multi-segment expert module consisting of \( M \) parallel local experts $\mathbf{\{E}_m^{(k)}(\cdot)\}_{m=1}^M$, each implemented by a lightweight MLP. These experts are intended to capture diverse, potentially complementary representations of the intra-segment features. Specifically, the generated hidden representation of the \( m \)-th expert and $k$-th frequency segment $\mathbf{H}^{(k)}_m$ is:
\begin{equation}
\mathbf{H}^{(k)}_m = \mathbf{E}_m^{(k)}\bigl( \mathbf{F}^{(k)} \bigr).
\end{equation}

When \( M=1 \), the output of the single expert is directly used. For \( M>1 \), the outputs of all experts are averaged to yield the representation of the $k$-th frequency segment $\mathbf{H}^{(k)}$:
\begin{equation}
\mathbf{H}^{(k)} =
\frac{1}{M} \sum_{m=1}^M \mathbf{H}^{(k)}_m.
\end{equation}

\( \mathbf{H}^{(k)} \) processes a single frequency band \( \mathbf{F}^{(k)} \) and integrates the output results of $M$ expert networks by averaging, thereby fully leveraging the collaborative advantages of multiple expert networks to significantly enhance overall performance. This design can balance expressiveness and computational efficiency, avoiding intra-segment gating while still allowing for ensemble effects among experts.
\subsubsection{Global Expert Module}

While segment-wise experts focus on local frequency bands, it is equally important to model global dependencies that span the entire frequency spectrum. To this end, we incorporate \( N \) global experts \( \{\mathbf{U}_i(\cdot)\}_{i=1}^N \), each processing the complete frequency spectrum signal \( \mathbf{F} \in \mathbb{C}^{d \times s_{\text{padded}}} \) to extract holistic representations:
\begin{equation}
\mathbf{H}^i_{\text{global}} = \mathbf{U}_i(\mathbf{F}).
\end{equation}

To adaptively weight the contributions of different global experts based on the input, 
we introduce a global gate $\sigma_{\text{global}}(\cdot)$, 
which is implemented as a simple linear layer followed by a softmax:

\begin{equation}
\lambda\ = \sigma_{\text{global}}\left( \mathbf{F} \right) \in \mathbb{R}^{N},
\end{equation}
where $\mathbf{F}$ is the input full-spectrum frequency representation, and $\lambda$ provides the normalized weights for all $N$ global experts. 
The final global representation $\mathbf{H}_{\text{global}}$ is then computed as a weighted sum of all the experts:

\begin{equation}
\mathbf{H}_{\text{global}} =
\sum_{i=1}^{N} \lambda_i \cdot \mathbf{H}^{\text{i}}_{\text{global},}
\end{equation}

\noindent where \( \mathbf{H}_{\text{global}} \in \mathbb{C}^{d \times s_{\text{padded}}} \). This mechanism allows the model to dynamically adjust which global experts to emphasize, providing flexibility to adapt to varying signal characteristics.

\subsubsection{Adaptive Gating Mechanism}

Beyond global expert fusion, we also introduce a segment-level gating mechanism to adaptively combine segment-wise representations. 
The original full frequency domain information $\mathbf{F}$ is fed into a segment gate \( \sigma_{\text{segment}}(\cdot) \), which directly produces normalized weights \( \beta\) for each segment:

\begin{equation}
\beta = \sigma_{\text{segment}}\left( \mathbf{F} \right) \in \mathbb{R}^{K}.
\end{equation}

The local representation is then computed as a weighted combination of segment outputs:
\begin{equation}
\mathbf{H}_{\text{local}} = \text{Concat} \left( \beta_1 \cdot \mathbf{H}^{(1)}, \dots, \beta_K \cdot \mathbf{H}^{(K)} \right),
\end{equation}

\noindent where  \( \mathbf{H}_{\text{local}} \in \mathbb{C}^{d \times s_{\text{padded}}} \) and 
$\beta_k$ provides the softmax-normalized importance weights for segment $k$, enabling the model to dynamically balance contributions from different frequency bands.

We have achieved global frequency spectrum representation $\mathbf{H}_{\text{global}}$ and fused representation from individual frequency segments $\mathbf{H}_{\text{local}}$. 
Then, we leverage a gate network $\sigma_{reweight}$ for global representation $\mathbf{H}_{\text{global}}$ to adaptively modulate the channel in the complete frequency spectrum.
Finally, to yield a comprehensive representation, we combine the original frequency-domain signal $\mathbf{F}$, the adaptively fused local representation $\mathbf{H}_{\text{local}}$, and the adaptively fused global representation $\mathbf{H}_{\text{global}}$:
\begin{equation}
\mathbf{H}^* =
\mathbf{F} + \mathbf{H}_{\text{local}} +
 \gamma \cdot \hat{\alpha}  \cdot \mathbf{H}_{\text{global}}.
\end{equation}

Here, \( \gamma \) is a learnable scalar obtained through the local and global two-path features, and $\hat{\alpha}$ denotes the adaptive gating weight. 

\subsection{Prediction Layer}

The final frequency-domain representation \( \mathbf{H}^* \) is transformed back into the time domain using the inverse FFT:
\begin{equation}
\mathbf{X}^* = \mathrm{iFFT}(\mathbf{H}^*).
\end{equation}

This reconstructed time-domain signal is passed through a fully connected classification layer and a softmax activation to produce class probabilities $\hat{y}$:
\begin{equation}
\hat{y} =
\text{Softmax}(W \cdot \mathbf{X}^* + b),
\end{equation}
where $W$ and $b$ represent the weight and bias of the linear layer.
This end-to-end pipeline enables the model to predict the class label based on frequency-aware representations.

\begin{table*}[t]
\centering
\setlength{\tabcolsep}{3.2pt}
\renewcommand{\arraystretch}{1}

\begin{tabular}{l|ccccccccccc}
\toprule
\textbf{Dataset} &
{\textit{\footnotesize DTWD}} &
{\textit{\footnotesize TapNet}} &
{\textit{\footnotesize ShapeNet}} &
{\textit{\footnotesize TNC}} &
{\textit{\footnotesize TS2Vec}} &
{\textit{\footnotesize InfoMin}} &
{\textit{\footnotesize InfoTS}} &
{\textit{\footnotesize AutoTCL}} &
{\textit{\footnotesize MPTSNet}} &
{\textit{\footnotesize FreRA}} &
{\textbf{\footnotesize Ours}} \\
\midrule

ArticularyWordRecognition & 98.7 & 98.7 & 98.7 & 97.3 & 98.7 & 91.3 & 98.7 & 98.3 & 97.7 & \underline{99.0} & \textbf{99.3} \\
AtrialFibrillation & 20.0 & 33.3 & 40.0 & 13.3 & 20.0 & 26.7 & 20.0 & 46.7 & \underline{53.3} & 46.7 & \textbf{73.3} \\
BasicMotions & \underline{97.5} & \textbf{100.0} & \textbf{100.0} & \underline{97.5} & \underline{97.5} & \textbf{100.0} & \underline{97.5} & \textbf{100.0} & \textbf{100.0} & \textbf{100.0} & \textbf{100.0} \\
CharacterTrajectories & 98.9 & \textbf{99.7} & 98.0 & 96.7 & \underline{99.5} & 99.0 & 97.4 & 97.6 & - & 99.1 & 99.2 \\
Cricket & \textbf{100.0} & 95.8 & \underline{98.6} & 95.8 & 97.2 & 95.8 & \underline{98.6} & \textbf{100.0} & 94.4 & \textbf{100.0} & 98.3 \\
DuckDuckGeese & 60.0 & 57.5 & \underline{72.5} & 46.0 & 68.0 & 70.0 & 54.0 & 70.0 & 68.0 & \textbf{76.0} & 68.0 \\
EigenWorms & 61.8 & 57.5 & 72.5 & 84.0 & 84.7 & 79.4 & 73.3 & \textbf{90.1} & - & \underline{86.3} & 55.7 \\
Epilepsy & 96.4 & 97.1 & \underline{98.7} & 95.7 & 96.4 & 92.0 & 97.1 & 97.8 & 97.1 & \textbf{99.3} & 87.0 \\
EthanolConcentration & 32.3 & 32.3 & 31.2 & \textbf{85.2} & 30.8 & 24.3 & 28.1 & 35.4 & \underline{43.3} & 32.3 & 33.5 \\
ERing & 13.3 & 13.3 & 13.3 & 29.7 & 87.4 & 90.4 & \underline{94.9} & 94.4 & 94.4 & 91.9 & \textbf{97.4} \\
FaceDetection & 52.9 & 55.6 & 60.2 & 53.6 & 50.1 & 56.0 & 53.4 & 58.1 & \underline{69.8} & 58.1 & \textbf{70.1} \\
FingerMovements & 53.0 & 53.0 & 58.9 & 47.0 & 48.0 & 50.0 & 63.0 & \underline{64.0} & \underline{64.0} & 61.0 & \textbf{68.0} \\
HandMovementDirection & 23.1 & 37.8 & 33.8 & 32.4 & 33.8 & 32.4 & 39.2 & 43.2 & \underline{63.5} & 51.4 & \textbf{68.9} \\
Handwriting & \textbf{60.7} & 35.7 & 45.1 & 24.9 & 51.5 & 56.9 & 45.2 & 38.4 & 34.4 & \underline{59.3} & 36.5 \\
Heartbeat & 71.7 & 75.1 & 75.6 & 74.6 & 68.3 & 73.7 & 72.2 & \underline{78.5} & 75.6 & \underline{78.5} & \textbf{79.0} \\
JapaneseVowels & 94.9 & 96.5 & 98.4 & 97.8 & 98.4 & 93.8 & 98.4 & 98.4 & \underline{98.6} & 96.5 & \textbf{99.2} \\
Libras & 87.2 & 85.0 & 85.6 & 81.7 & 86.7 & 80.0 & 88.3 & 83.3 & 87.2 & \underline{91.1} & \textbf{91.7} \\
LSST & 55.1 & 56.8 & 59.0 & \underline{59.5} & 53.7 & 47.3 & 59.1 & 55.4 & \textbf{60.4} & 49.4 & 39.0 \\
MotorImagery & 50.0 & 59.0 & 61.0 & 50.0 & 51.0 & 53.0 & 63.0 & 57.0 & \textbf{65.0} & 55.0 & \underline{64.0} \\
NATOPS & 88.3 & 93.9 & 88.3 & 91.1 & 92.8 & 82.2 & 93.3 & \underline{94.4} & \underline{94.4} & 90.0 & \textbf{97.2} \\
PEMS-SF & 71.1 & 75.1 & 75.1 & 69.9 & 68.2 & 69.9 & 75.1 & 83.8 & \textbf{94.2} & 74.6 & \underline{93.6} \\
PenDigits & 97.7 & 98.0 & 97.7 & 97.9 & \underline{98.9} & 97.0 & \textbf{99.0} & 98.4 & \underline{98.9} & 97.3 & \underline{98.9} \\
PhonemeSpectra & 15.1 & 17.5 & \textbf{29.8} & 20.7 & 23.3 & 24.0 & 24.9 & 21.8 & 14.4 & 27.4 & 14.4 \\
RacketSports & 80.3 & 86.8 & 88.2 & 77.6 & 85.5 & 82.2 & 85.5 & \textbf{91.4} & 87.5 & 88.8 & \underline{90.8} \\
SelfRegulationSCP1 & 77.5 & 65.2 & 78.2 & 79.9 & 81.2 & 86.7 & 87.4 & 89.1 & \textbf{92.8} & \underline{90.8} & \textbf{92.8} \\
SelfRegulationSCP2 & 53.9 & 55.0 & 57.8 & 55.0 & 57.8 & \underline{62.0} & 57.8 & 57.8 & 57.2 & \textbf{62.2} & 59.4 \\
SpokenArabicDigits & 96.3 & 98.3 & 97.5 & 93.4 & 93.2 & 98.1 & 94.7 & 92.5 & \underline{99.5} & 98.4 & \textbf{99.8} \\
StandWalkJump & 20.0 & 40.0 & 53.3 & 40.0 & 46.7 & 33.3 & 46.7 & 53.3 & 53.3 & \textbf{66.7} & \underline{60.0} \\
UWaveGestureLibrary & \underline{90.3} & 89.4 & \textbf{90.6} & 75.9 & 88.4 & 87.2 & 88.4 & 89.3 & 88.1 & 90.0 & \textbf{90.6} \\
InsectWingbeat & - & 20.8 & 25.0 & 46.9 & 46.6 & 44.3 & 47.0 & \underline{48.8} & - & 46.2 & \textbf{56.8} \\

\midrule
\textbf{Top-1 count ↑} & 2 & 2 & 3 & 1 & 0 & 1 & 1 & 4 & 5& \underline{6} & \textbf{15} \\
\textbf{Avg. Acc. (\%) ↑} & 63.9 & 66.0 & 69.4 & 67.0 & 70.1 & 69.3 & 71.4 & 74.2 & 68.2 & \underline{75.4} & \textbf{76.1} \\
\textbf{Avg. Rank ↓} &7.6 	&6.3 	&5.0 	&8.2 	&6.6 	&7.5 	&5.4 	&4.1 	&4.7 	&\underline{3.8} 	&\textbf{3.2} 
 \\
\bottomrule
\end{tabular}

\caption{Overall experimental comparison results on 30 datasets. 
The best results are highlighted in \textbf{bold}, and the second-best results are marked with an \underline{underline}. ↑ indicates higher is better, and ↓ indicates lower is better. 
}
\label{tab:large-comparison}
\end{table*}

\subsection{Optimization Objective Function}

\begin{table*}[t]
\small
\setlength{\tabcolsep}{2pt} 
\renewcommand{\arraystretch}{1}

\begin{tabular}{l|cccccccccccc}
\toprule
\multirow{2}{*}{\textbf{Dataset}}      & LSTNet  & LSSL     & FEDf.   & Flowf.  & SCINet  & Dlinear & PatchTST & MICN    & TimesNet & Crossf.  & M.TCN   & \multirow{2}{*}{\textbf{Ours}} \\ 
                       & \textit{\scriptsize SIG'18}  & \textit{\scriptsize ICLR'22}  & \textit{\scriptsize ICLR'22} & \textit{\scriptsize ICLR'22} & \textit{\scriptsize NIPS'22} & \textit{\scriptsize AAAI'23} & \textit{\scriptsize ICLR'23}  & \textit{\scriptsize ICLR'23} & \textit{\scriptsize ICLR'23}  & \textit{\scriptsize ICLR'23}  & \textit{\scriptsize ICLR'24} &      \\ \midrule

EthanolConcentration           & \textbf{39.9}    & 31.1     & 31.2    & 33.8    & 34.4    & 36.2    & 32.8     & 35.3    & 35.7     & \underline{38.0}     & 36.3    & 33.5 \\
FaceDetection            & 65.7    & 66.7     & 66.0    & 67.6    & 68.9    & 68.0    & 68.3     & 65.2    & 68.6     & 68.7     & \textbf{70.8}    & \underline{70.1} \\
Handwriting            & 25.8    & 24.6     & 28.0    & \underline{33.8}    & 23.6    & 27.0    & 29.6     & 25.5    & 32.1     & 28.8     & 30.6    & \textbf{36.5} \\
Heartbeat              & 77.1    & 72.7     & 73.7    & 77.6    & 77.5    & 75.1    & 74.9     & 74.7    & \underline{78.0}     & 77.6     & 77.2    & \textbf{79.0} \\
JapaneseVowels         & 98.1    & 98.4     & 98.4    & 98.9    & 96.0    & 96.2    & 97.5     & 94.6    & 98.4     & \underline{99.1}     & 98.8    & \textbf{99.2} \\
PEMS-SF                & 86.7    & 86.1     & 80.9    & 86.0    & 83.8    & 75.1    & 89.3     & 85.5    & \underline{89.6 }    & 85.9     & 89.1    & \textbf{93.6} \\
SelfReglationSCP1            & 84.0    & 90.8     & 88.7    & 92.5    & 92.5    & 87.3    & 90.7     & 86.0    & 91.8     & 92.1     & \textbf{93.4}    & \underline{92.8} \\
SelfReglationSCP2            & 52.8    & 52.2     & 54.4    & 56.1    & 57.2    & 50.5    & 57.8     & 53.6    & 57.2     & 58.3     & \textbf{60.3}    & \underline{59.4} \\
SpokenArabicDigits       & \textbf{100}   & \textbf{100}    & \textbf{100}   & 98.8    & 98.1    & 81.4    & 98.3     & 97.1    & 99.0     & 97.9     & 98.7    & \underline{99.8} \\
UWaveGestureLibrary       & \underline{87.8}    & 85.9     & 85.3    & 86.6    & 85.1    & 82.1    & 85.8     & 82.8    & 85.3     & 85.3     & 86.7    & \textbf{90.6} \\ 
\midrule
\textbf{Top-1 count ↑} &2 &1 &1 &0  &0  &0  &0  &0  &0  &0  &3 &\textbf{5} \\
\textbf{Avg. Acc. (\%) ↑} & 71.8    & 70.9     & 70.7    & 73.2    & 71.7    & 67.9    & 72.5     & 70.0    & 73.6     & 73.2     & \underline{74.2}    & \textbf{75.5} \\
\textbf{Avg. Rank ↓} & 6.6     & 7.9      & 8.0     & 5.1     & 7.5     & 9.5     & 6.8      & 10.1    & 4.5      & 5.0      & \underline{3.4}     & \textbf{2.4} \\ 
\bottomrule

\end{tabular}
\centering
\caption{Experimental comparison on 10 UEA datasets. The best results are highlighted in \textbf{bold}, and the second-best results are marked with an \underline{underline}. ↑ indicates higher is better, and ↓ indicates lower is better.  
}
\label{tab:comparison}
\end{table*}

In MTSC tasks, data distributions are complex, class imbalance is common, and datasets differ significantly. Cross Entropy-Loss, which is widely used in prior work, struggles with class imbalance and fails to distinguish between easy and hard samples, leading to poor handling of minority classes and complex instances. To address this, we propose adopting Polynomial Loss~\cite{leng2022polyloss} for time series classification.

Polynomial Loss was initially proposed to improve optimization and calibration performance in general classification tasks.  
However, the original Polynomial Loss introduces high-order terms and multiple hyperparameters, which may complicate the optimization process in MTSC scenarios.
In this case, we simplify and customize it into an improved loss function called \textbf{P-Loss}.

We propose P-Loss to optimize the original formula by retaining only the core second-order adjustment term while preserving the stability and robustness advantages. The second-order formulation is adopted because it provides nonlinear gradient corrections with negligible computational overhead, effectively optimizing the classification boundaries for multi-channel data without introducing the noise associated with higher-order terms. Specifically, given \( N \) samples, where \( \hat{y}_i \) represents the predicted probability of the true class for the \( i \)-th sample, P-Loss is defined as:

\begin{equation}
\mathcal{L}_P = -\frac{1}{N} \sum_{i=1}^N \log \hat{y}_i + \lambda_P \cdot \frac{1}{N} \sum_{i=1}^N (1 - \hat{y}_i)^2
\end{equation}

The first term is the negative log-likelihood, maximizing the predicted probability \(\hat{y}_i\), part of the standard cross-entropy loss. The second term is a second-order adjustment that penalizes deviations from the true label, with \(\lambda_P\) controlling its weight.

We also incorporate mixup with a simple adaptive adjustment. If the loss stagnates after several training rounds, the mixing ratio is dynamically reduced to bring the mixed samples closer to the original ones, improving generalization without harming baseline performance. Combined with P-Loss, this design simplifies optimization while preserving model stability and robustness, ultimately demonstrating stronger generalization on diverse MTSC benchmarks.

\subsection{Computational Complexity Analysis}

Most operations in FreSH, except for the frequency-domain transformation, are linear in the sequence length, which makes the framework computationally efficient. The overall computational complexity of a single forward pass through FreSH can be expressed as $O(d n \log n + mhn + ghn)$, where $d$ is the number of channels (variables) in the multivariate time series, $n$ is the frequency-domain vector length of the input, $m$ is the number of experts per frequency segment, $g$ is the number of global experts, and $h$ is the hidden dimension of each expert. The $O(d n\log n)$ term comes from the FFT and inverse FFT transformations applied across $d$ channels, which dominate for large $n$. The segment-wise expert networks contribute $O(mhn)$, as each of the $k$ segments independently applies $m$ small MLPs on $n/k$ elements. Similarly, the global experts contribute $O(ghn)$ by processing the full $n$-dimensional vector through $g$ MLPs. Since $d$, $m$, $g$, and $h$ are small constants compared to $n$ in practice, the computational complexity grows approximately as $O(d n \log n)$, making the model efficient and scalable for long multivariate time series.

\section{Experiments}

\subsection{Experimental Setups}
We conduct experimental comparison on the UEA MTSC benchmarks~\cite{bagnall2018uea}, which span applications such as human activity recognition, speech processing, medical EEG, and audio analysis. The datasets differ substantially in sequence length, dimensionality, and train/test sizes, enabling a robust assessment of generalization.

To systematically evaluate the effectiveness of our proposed FreSH, we conduct comprehensive comparisons against a wide range of state-of-the-art baselines.

On one hand, our main experiments focus on comparisons with dedicated multivariate time series classification methods. Specifically, we consider both traditional and recent MTSC approaches, including DTWD, ShapeNet~\cite{li2021shapenet}, TapNet~\cite{zhang2020tapnet}, TNC ~\cite{tonekaboni2021unsupervised}, TS2Vec~\cite{yue2022ts2vec}, InfoMin~\cite{tian2020makes}, InfoTS~\cite{luo2023time}, AutoTCL~\cite{zheng2023auto}, MPTSNet~\cite{mu2025mptsnet}, and FreRA~\cite{tian2025frera}. These methods are evaluated on 30 UEA multivariate time series datasets, providing a thorough and fair comparison with existing MTSC models.

On the other hand, to comprehensively position our proposed FreSH, we compare FreSH with representative time series representation models, including LSTNet~\cite{lai2018modeling}, LSSL~\cite{gu2021efficiently}, TimesNet~\cite{wu2022timesnet}, PatchTST~\cite{nie2022time}, FlowFormer~\cite{wu2022flowformer}, FEDformer~\cite{zhou2022fedformer}, SCINet~\cite{liu2022scinet}, DLinear~\cite{zeng2023transformers}, Crossformer~\cite{zhang2023crossformer}, MICN~\cite{wang2023micn}, and ModernTCN~\cite{luo2024moderntcn}. 
Following the general setting~\cite{mu2025mptsnet}, we conduct the comparison on 10 UEA datasets.
All experiments are conducted on an NVIDIA A100. We report average accuracy (Avg. Acc.), average rank (Avg. Rank), and the number of best-accuracy datasets (Top-1 count) to evaluate MTSC performance and enable fair comparisons with baseline methods.

\subsection{Experimental Results}
Table \ref{tab:large-comparison} presents an extensive empirical evaluation of FreSH against 10 state-of-the-art Multivariate Time Series Classification (MTSC) baselines across a diverse collection of 30 UEA datasets. The quantitative results demonstrate the overwhelming superiority and robustness of our proposed method.

In terms of overall performance across all 30 datasets, FreSH establishes a new state-of-the-art benchmark. It achieves the highest average accuracy of 76.1\% and the best (lowest) average rank of 3.4 among all compared methods. When compared to the second-best performer, FreRA, FreSH not only improves the average accuracy by 0.7\% (76.1\% vs. 75.4\%) but also demonstrates greater consistency, surpassing FreRA by a margin of 0.6 in average ranking (3.2 vs. 3.8). This indicates that FreSH maintains high performance stability across varying data domains.

A deeper analysis of the specific 10 MTSC datasets (Table \ref{tab:comparison}) further highlights the model's adaptability. In this challenging subset, the performance gap between FreSH and existing methods becomes even more pronounced. Compared to the strong baseline ModernTCN, which ranks second in both metrics, FreSH delivers a substantial improvement: it boosts the average accuracy by 1.3\% and advances the average ranking by a full position (1.0).

Beyond the average metrics, the ranking distribution across datasets further confirms the core advantages of our method. In the evaluation of the complete UEA dataset, FreSH achieved the highest number of Top-1 results, significantly surpassing the next best model. Furthermore, in the evaluation setting across 10 datasets, FreSH achieved best accuracy on 5 datasets and second best accuracy on 4 datasets. 

Overall, this outstanding performance across multiple metrics and settings demonstrates that FreSH is an excellent model capable of handling complex temporal patterns, significantly outperforming both traditional and recently proposed MTSC models, with remarkable adaptability and effectiveness.

\subsection{Ablation Study}
\begin{table}[htbp]
\label{tab:Ablation}
\setlength{\tabcolsep}{25pt} 
\renewcommand{\arraystretch}{1} 
\begin{tabular}{l|c}
\toprule
Model                   & Avg. Acc. (\%)        \\ 
\midrule
w/o-GlobalExperts  & 73.3         \\
w/o-SegmentsExperts & 73.1          \\
w/o-P-loss          & 72.9          \\
w/o-Mixup          & 74.6          \\
w/o-$\beta$\        &73.4           \\
w/o-$\lambda$\       &74.3           \\

\midrule
\textbf{FreSH}              & \textbf{76.1} \\
\bottomrule
\end{tabular}
\caption{Ablation experiments on UEA datasets.}
\end{table}

We conduct a comprehensive ablation study to investigate the contributions of the key components in our model. 
\begin{itemize}[leftmargin=*]
    \item \textbf{w/o-GlobalExperts}: Remove the global experts and only rely on segment-wise experts for feature extraction.
    \item \textbf{w/o-SegmentExperts}: Remove the segment-wise experts, relying solely on global experts.
    \item \textbf{w/o-P-Loss}: Replace the P-loss with cross-entropy loss.
    \item \textbf{w/o-Mixup}: Disable the adaptive mixup data augmentation.
    \item \textbf{w/o-$\beta$}: Disable adaptive gating fusion for Segment experts.
    \item \textbf{w/o-$\lambda$}: Disable adaptive gating fusion for Global experts.

\end{itemize}

The ablation study confirms that all components of the FreSH framework are critical to its performance. The largest drop in accuracy occurs when removing P-Loss 
(76.1\% to 72.9\%),
a strong indicator of its central role in achieving robust optimization for imbalanced and difficult samples. 
The hierarchical multi-expert framework is also essential, as removing either the global experts (76.1\% to 73.3\%)
or the segment experts 
(76.1\% to 73.1\%)
significantly degrades performance, validating the synergy between local feature analysis and global information integration. 
While the mixup augmentation also contributes, its removal leads to a comparatively smaller drop 
(76.1\% to 74.6\%),
showing it is a valuable but supplementary component. 
Under the premise of retaining the linear expert, ablation experiments were conducted on the adaptive gated fusion of local ($\beta$) and global ($\lambda$) experts. Although the removal of this fusion module led to a decrease in average accuracy (2.7\% and 1.8\%), its performance still significantly outperformed the setting where all experts were removed simultaneously, with dual comparisons validating that the linear expert provides a stable baseline and the adaptive fusion further achieves effective collaboration among experts.

\subsection{Efficiency Comparison}

To assess the prediction precision and computational efficiency in realistic settings, we benchmark all models on a real-world vibration dataset. We report model parameter volume, average batch latency, total test time, and classification accuracy to characterize the trade-off between computational cost and predictive performance.
\begin{table}[htbp]
\centering
\setlength{\tabcolsep}{1pt}
\renewcommand{\arraystretch}{1}
\begin{tabular}{l|cccc}
\toprule
Model & \#Params$\downarrow$ & Batch (ms)$\downarrow$ & Test (s)$\downarrow$ & Acc. (\%)$\uparrow$ \\
\midrule
Autoformer    & 933,507      & 23.5 &  8.518         &    85.06   \\
TimesNet      & 653,651     & 32.5  & 11.754   & 84.16 \\
Informer      & 1,139,338   & 9.2   & 3.314    & 90.10 \\
FEDformer     &1,531,402    & 48.1      & 17.407 &91.10 \\         
PatchTST      & 466,947     & 2.8   & 0.976    & \underline{94.34} \\
DLinear       & 3,927,003   & \textbf{0.66} & \textbf{0.187}    & 44.89 \\
Crossformer   & 2,843,059   & 25.3  & 1.468    & 90.30 \\
Transformer   & 936,195     & 63.9  & 23.131   & 89.30 \\
LightTS       & 2,026,909   & \underline{1.2}   & 0.372    & 88.27 \\

MPTSNet       & 88,181,371  & 101.4 & 123.325  & 84.51 \\
FreRA         & 516,299,712          & 8.8    & 0.966 & 81.45 \\
\midrule
FreSH         & \textbf{54,243} & \underline{1.2} & \underline{0.344} & \textbf{94.37} \\
\bottomrule
\end{tabular}
\caption{Efficiency comparison on a real-world vibration dataset. Certain models adopt a classification adaptation scheme identical to FreSH.
}
\label{tab:vibration-efficiency}
\end{table}

Table~\ref{tab:vibration-efficiency} shows large efficiency gaps. DLinear is extremely fast (0.66 ms/batch; 0.187 s total) but underfits vibration signals, yielding low accuracy (44.89\%). In contrast, TimesNet, Crossformer, and other Transformer-based models achieve competitive accuracy at much higher cost (25.3–63.9 ms/batch). MPTSNet further demonstrates diminishing returns: despite strong capacity, its $>$ 88M parameters lead to prohibitive latency (101.4 ms/batch; $>$ 123 s total), limiting practicality.
By contrast, our proposed FreSH achieves the best balance between performance and efficiency. With only 54,243 parameters, it remains fast (1.2 ms/batch; 0.344 s total) while achieving 94.37\% accuracy. 

These results indicate that FreSH effectively reconciles efficiency and accuracy, making it well-suited for real-world vibration analysis where both are critical.

\subsection{Hyper-Parameter Analysis}

\begin{figure}[!t]
\centering
\includegraphics[width=1\columnwidth]{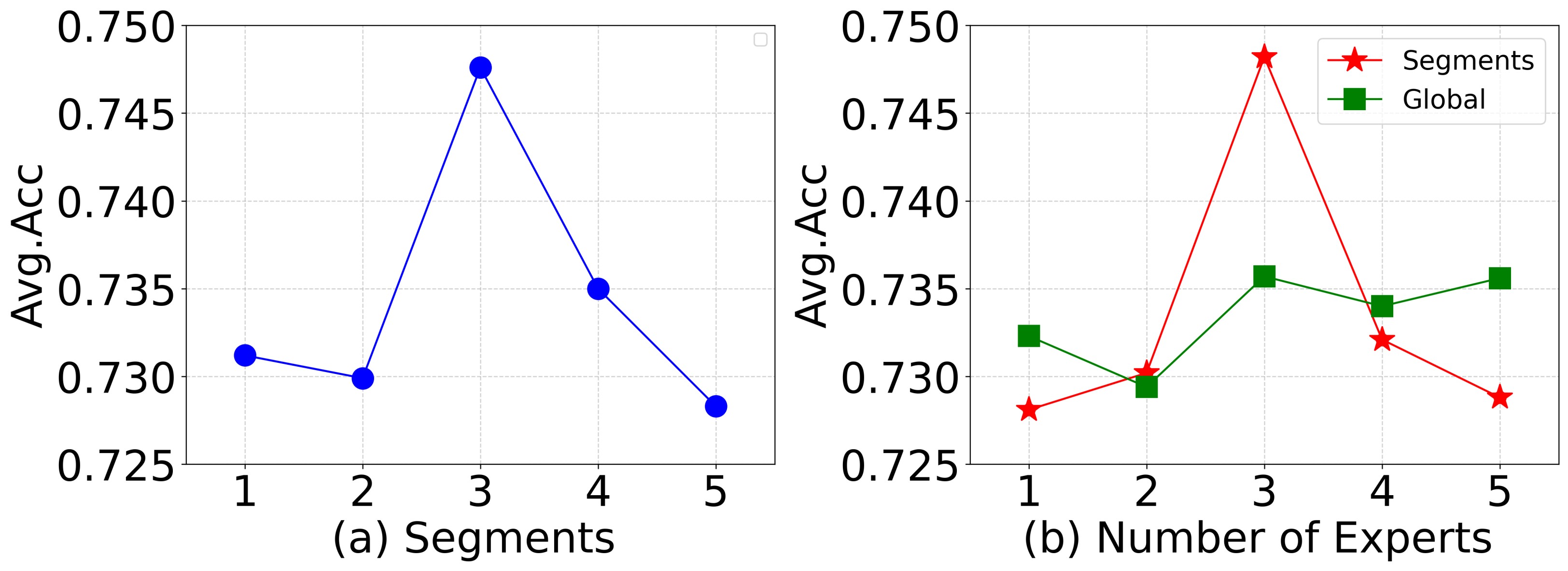} 
\caption{Experiments on 30 UEA datasets to evaluate different structural configurations, we report the average accuracy. }
\label{fig3}
\end{figure}

We conduct quantitative experiments to evaluate the contributions of different components of FreSH. 
Figure~\ref{fig3} shows average accuracy on 30 UEA datasets under various hyperparameter settings. Figure~\ref{fig3} (a) tests the number of frequency segments. Figure~\ref{fig3} (b)  examines experts per segment and analyzes global experts.

Experimental results shown in Figure~\ref{fig3} highlight the critical role of structural design in our framework. Specifically, segmenting the frequency spectrum into 3 bands strikes a balance between capturing fine-grained local patterns and preserving sufficient global context. Likewise, assigning 3 experts enables diverse feature extraction within each band without introducing unnecessary redundancy. 

In contrast, overly fine segmentation fragments the spectrum and dilutes useful information, while too many experts per segment increases the risk of overfitting and leads to unstable representations. 

These findings confirm that carefully calibrating the segmentation granularity and expert allocation is essential for effectively modeling both local and global dependencies in multivariate time series classification.

\section{Related Work}
\subsection{Time-Domain MTSC Methods}
A substantial portion of multivariate time series classification research has concentrated on modeling signals directly in the time domain. Notably, the MC-DCNN~\cite{zheng2014time} applies one-dimensional convolutions to capture inter-variable relations, pairing them with fully connected layers for classification purposes. Building on this, the Multiscale Convolutional Neural Network (MSCNN) designs convolution kernels of multiple sizes to extract multiscale features~\cite{cui2016multi}. Hybrid architectures have also been explored, such as LSTM-FCN~\cite{karim2017lstm}, which leverages LSTM layers for short- and long-term dependency capture while CNN layers extract salient time series patterns. Its enhanced version MLSTM-FCN~\cite{karim2019multivariate} further refines this combination for performance gains. More recent developments like TimesNet~\cite{wu2022timesnet} disentangle complex temporal variations into intra- and inter-period components for improved local and global feature modeling, whereas MS-GNet~\cite{cai2024msgnet} integrates graph convolution for inter-series correlation alongside multi-head attention for intra-series feature learning. Likewise, PatchTST~\cite{nie2022time} partitions sequences into local patches to capture hierarchical patterns, while Autoformer~\cite{wu2021autoformer}  proposes a novel auto-correlation mechanism that captures long-range dependencies by calculating the periodic similarity of time series.

Compared with this line of methods, our FreSH's hierarchical multi-expert architecture allows us to specifically extract the crucial frequency characteristics, achieving superior computational efficiency and adaptability. 

\subsection{Frequency-Domain MTSC Methods}
In parallel, a growing body of work has sought to harness frequency-domain representations for MTSC, leveraging spectral analysis to uncover new optimization pathways. For instance, the Frequency-improved Legendre Memory Model~\cite{zhou2022film} augments Legendre memory structures with spectral components, markedly improving long-term sequence classification. CrossFormer~\cite{zhang2023crossformer} unifies frequency-domain decomposition with Transformer architectures to jointly refine local and global feature extraction, while MPTSNet~\cite{mu2025mptsnet} explicitly utilizes amplitude information to detect salient periodicities for rapid feature localization. Other methods bypass the time domain entirely: FreTS~\cite{yi2023frequency} demonstrates the compactness of spectral information and constructs a frequency-domain MLP to achieve state-of-the-art performance, and FreRA~\cite{tian2025frera} proposes a parameterized augmentation strategy with time-frequency consistency constraints for robust training.

Existing frequency-domain methods typically perform a single, global analysis of the entire spectrum, which overlooks the unique information contained in different frequency bands. 
Our FreSH addresses this by segmenting the spectrum and using an adaptive expert system to process each segment individually, enabling a fine-grained and specialized analysis of each frequency band.

\section{Conclusion}
To advance multivariate time series classification, we propose FreSH, which combines frequency-domain analysis with an adaptive expert system. 

FreSH transforms time series into the frequency domain and applies spectral segmentation: segment experts capture band-specific features, global experts model full-spectrum context, and adaptive gating fuses them into a more balanced representation.

FreSH further introduces P-Loss to address the limitations of cross-entropy and focal loss, and incorporates an adaptive mixup strategy to improve robustness and generalization. 

Extensive experiments on a broad range of UEA benchmarks and a real-world vibration dataset demonstrate that FreSH consistently achieves superior or competitive accuracy compared to state-of-the-art methods, while maintaining high computational efficiency and a compact parameter footprint. The strong performance across diverse datasets underscores the framework's practical applicability and robustness, highlighting its potential for real-world deployment scenarios where both accuracy and efficiency are essential.

\section*{Acknowledgements}
This work was supported by Jilin Province Industrial Key
Core Technology Tackling Project (20230201085GX). Zijian Zhang is supported by the China Postdoctoral Science Foundation (2025M771587) and the Open Funding Programs of State Key Laboratory of AI Safety (2025-09).
Hao Miao is supported by SCRI, The Hong Kong Polytechnic University (No. Q-CDDG).
Qingliang Li is supported by the National Natural Science Foundation of China(42575159, 42275155, 62206028).

\bibliographystyle{named}

\bibliography{ijcai26}

\end{document}